\documentclass[conference,compsoc]{IEEEtran}
\ifCLASSOPTIONcompsoc
  \usepackage[nocompress]{cite}
\else
  \usepackage{cite}
\fi

\usepackage{times}
\usepackage{graphicx}
\usepackage{subfigure}
\graphicspath{{fig2/}}
\usepackage{epstopdf}		% SHOULD BE after graphicx!
\usepackage[cmex10]{amsmath}
\usepackage{longtable}
\usepackage{pdfpages}
\usepackage{bbding}
\usepackage{pifont}
\usepackage{wasysym}
\usepackage{amssymb}
\usepackage{listings}
\usepackage{url}
\usepackage{color}
\usepackage{setspace}
\usepackage{float}
\usepackage{multirow}

\begin{document}
\title{
An Ensemble Approach toward Automated Variable Selection \\  for  Network Anomaly Detection
}


%\author{\IEEEauthorblockN{AAA, (some more)}
%\IEEEauthorblockA{Computer Science Department\\
%Texas A\&M University\\
%Commerce, TX 75428\\
%Email: AAA@tamuc.edu}
%\and
%\IEEEauthorblockN{Authors Name/s per 2nd Affiliation (Author)}
%\IEEEauthorblockA{line 1 (of Affiliation): dept. name of organization\\
%line 2: name of organization, acronyms acceptable\\
%line 3: City, Country\\
%line 4: Email: name@xyz.com}
%}
\author{Makiya Nakashima\IEEEauthorrefmark{1}\, Alex Sim\IEEEauthorrefmark{2}\, Youngsoo Kim\IEEEauthorrefmark{3}\, Jonghyun Kim\IEEEauthorrefmark{3}\, Jinoh Kim\IEEEauthorrefmark{1}
\IEEEauthorblockA{\\Texas A\&M University, Commerce, TX 75428, USA \IEEEauthorrefmark{1} \\
Lawrence Berkeley National Laboratory, Berkeley, CA 94720, USA \IEEEauthorrefmark{2} \\
ETRI, Daejeon, 34129, Korea \IEEEauthorrefmark{3} \\
% Email: mnakashima@leomail.tamuc.edu}}
 Email: jinoh.kim.tamuc.edu
}}

\maketitle

\begin{abstract}
%Automating the process of variable selection is essential since it requires laborious efforts with intensive analysis otherwise.
%However, such an automation is highly challenging due to several reasons, including the difficulty to determine the termination condition to stop the iterative process of the selection method.
%Another important challenge would be that  there  is  no  sole  winner  that  outperforms  other selection  methods  at  all  occasions,  and  it  is  impossible  to predict  such  a  winner  in  the  training  time  even  if  there exists. 
%In this paper, we focus on the problem of identifying competitive  sets  of  features  as  an  initial  step  to  enable automated variable selection for network anomaly detection.
%We  present  an  ensemble  method  that employs  multiple  selection  techniques  to  see  the  potential  power  of  the  incorporation  of  identified  features  by individual  selection  techniques through the  heuristic functions defined in this study. 
%Our experimental results with two public traffic datasets show that two heuristics work well consistently identifying competitive subsets of features yielding the approximate performance to the one with the entire features.

While variable  selection  is  essential to optimize the learning complexity by prioritizing  features, automating the selection process is preferred since it requires  laborious efforts with intensive analysis otherwise.
However, it is not an easy task to enable the automation due to several reasons.
First, selection techniques often need a condition to terminate the reduction process, for example, by using a threshold or the number of features to stop, and searching an adequate stopping condition is highly challenging.
Second, it is uncertain that the reduced variable set would work well;   
our preliminary experimental result shows that well-known  selection  techniques produce different sets of variables as a result of reduction (even  with  the  same  termination  condition), and it is hard to estimate which of them would work the best in future testing.
In this paper, we demonstrate  the  potential  power  of  our approach to the automation of selection process that incorporates well-known selection methods  identifying important variables. %, using three heuristics  based on set theory.
%Our approach to the automation of selection process is to incorporate well-known selection methods  identifying important variables.
%In this paper, we demonstrate  the  potential  power  of  our ensemble method that incorporates multiple selection techniques using three heuristics  based on set theory \fix{to ?}. %, while we leave the determination of stopping condition as the next step towards the complete automation.
Our experimental results with two public network traffic data (UNSW-NB15 and IDS2017) show that our proposed method identifies a small number of core variables, with which it is possible to approximate the performance to the one with the entire variables. %  with a reduced computational complexity.
\end{abstract}

%\begin{IEEEkeywords}
%component; formatting; style; styling;
%\end{IEEEkeywords}

% \fi
\IEEEpeerreviewmaketitle

% Main text
\section{Introduction} 
	\label{sec:intro}

Rapid advance in the machine learning industry has led to the active adoption of the learning techniques for a diverse range of applications.
The network security area is not an exception and machine learning is widely employed to analyze network traffic.
For example, clustering has been utilized to understand temporal variation patterns, to detect network anomalies, and to infer the class of individual connections for the purpose of labeling.
Various classification techniques such as Random Forest (RF), Support Vector Machine (SVM), and deep neural networks (DNNs) have also been employed mainly for detecting intrusive and/or anomalous events by analyzing network traffic~\cite{kwon2017survey,buczak2015survey}.

\begin{figure}[!tb]
\centering
\includegraphics[width=.75\columnwidth]{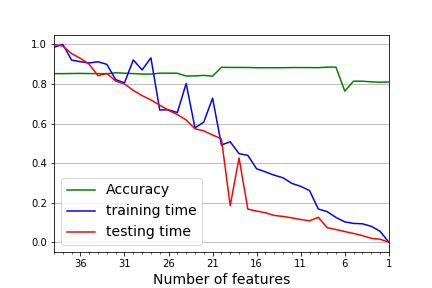}
\caption{
Accuracy and timing over the number of features (classifier=LR, dataset=UNSW-NB15): Training and testing times decrease almost linearly over the number of features, while accuracy shows  a flat pattern showing better performance with 7--20 features than the use of the entire features.
}
\label{fig:acc_time}
\end{figure}

A large body of the previous studies simply utilized the entire variables\footnote{We interchangeably use variables and features throughout the paper.} provisioned in the data files.
However, it is not always true that relying on all the variables leads to the best performance (e.g. detection rate); that is, reducing a degree of redundancy may result in even better result. 
In addition, an invariant is that relying on a greater number of variables should impose an increasing complexity, thus requiring a greater amount of time and resources for analyzing. %  for training and testing.
Fig.~\ref{fig:acc_time} shows an example of the detection accuracy and the (normalized) time taken for training and testing over the number of features with the Logistic Regression classifier against UNSW-NB15\footnote{We  utilize the popular UNSW-NB15 dataset since it has been collected in 2015. The description of the dataset employed in this paper will be provided in Section~\ref{sec:eval}.}
From the figure, training and testing times decrease almost linearly with a smaller number of features.
Interestingly, accuracy goes down eventually with smaller number of features ($\leq 6$), but it does not show the best performance with the entire 39 features (numeric only); rather, we can see better performance with 7--20 features.

Variable selection is a technique identifying relatively more important features by eliminating less essential variables. % that would be redundant or inessential.
Alternatively, dimensionality reduction tools such as Principal Component Analysis (PCA) and t-Distributed Stochastic Neighbor Embedding (t-SNE) project the given variables into a lower dimensional space, but the main difference is that they do not preserve the definitions of the individual features while variable selection does.   
Variable selection has been an active research topic for improving learning performance over a decade~\cite{li2018feature,alelyani2018feature}.
%However, the actual use of a selection technique is {\em not} straightforward even with laborious efforts by experts as discussed next.
%Hence, automating the selection process is much more challenging despite its large demand. 
While variable  selection  is  essential to optimize the learning complexity by identifying superior features, automating the selection process is desired since it requires  laborious efforts with intensive analysis otherwise.
However, it is {\em not} an easy task to enable the automation due to several reasons, as follows.

One of the challenges is that variable selection techniques often need to define a ``good'' termination condition to stop the elimination of features in its iterative process.
%the result of which leads to an approximated (or even better) performance in analysis compared to the use of the entire feature space. 
%For instance, the UNSW-NB15 dataset~\ref{} consists of XX number of features.
%A selection method eliminates less important features one by one at each iteration until it meets the final number of features specified.
Many feature selection methods rely on a threshold or the final number of features to reduce.
For instance, Sequential Feature Selection (SFS) eliminates less important features one by one at each iteration until it meets the final number of features specified.
A critical problem here is  how to determine the final number of features or the value for the threshold to stop.
%In this paper, we present a method to identify a subset of variables that would be essential to secure the learning performance.  
%The goal of this study is to answer the question above to identify a subset of variables that would be essential to secure the learning performance. \fix{It needs to address the critical problem above to determine the feature threshold in the goal, not only the feature selection. That critical problem is the main challenge in this paper, it seems, in addition to the better feature selection with ensemble method.}
Since the variable selection is performed in the {\em training} phase with some learning samples, which should be disjoint to the actual data used in the {\em testing} phase, the resources to help determine  appropriate stopping conditions should be limited to the information that can be available in the training time.

Another important challenge %for the effective variable selection 
would be how to discover an optimal subset of features that approximates to the performance obtained with the entire features.
%In fact, the variable selection is performed in the {\em training} phase with some learning samples, which should be disjoint to the actual data used in the {\em testing} phase.
%The selected variables will be working well with the testing data {\em only if} the selection is successful. 
One of our initial observations is that individual selection techniques employed in our experiments result in different feature sets, even with the same termination condition.
Another interesting observation is that the actual testing performance shows a degree of fluctuations over the iterative reduction process even with a single selection method, making it further difficult to estimate the termination point.
Moreover, the performance in the training phase does not guarantee the performance in the testing time for individual selection methods; for example, a selection method shows an outstanding performance but could reveal a relatively poor performance compared to other methods in the testing time.
%The goal of this study is to  identify an optimal subset of features in the training time without laborious human efforts. 
%In this paper, we take an initial step to enable automated variable selection for network anomaly detection through empirical experiments. 
In addition, there is {\em no} sole winner that  outperforms other selection methods at all occasions, and it is impossible to predict such a winner in the training time even if there exists.  
%In this paper, we focus on the problem of identifying competitive sets of features as an initial step to enable automated variable selection for network anomaly detection. 

Our approach to the automation of selection process is to incorporate well-known selection methods to make a safer identification of important variables.
In this paper, we present an ensemble method that employs multiple selection techniques to see the potential power of the incorporation of identified features by individual selection techniques,
while we leave the determination of stopping condition as the next step towards the automation of selection process.
As mentioned,
individual selection techniques produce slightly different sets at each iteration. 
%We combine the independent results by selection methods using three heuristics defined based on set theory: (1) {\em intersection} taking the features everyone agrees upon as a candidate feature set, (2) {\em union} taking the ones anyone suggests, and (3) {\em quorum} selecting ones a majority of the participants (selection methods) consent on.
To combine the  results by individual selection schemes, we establish three heuristic methods defined based on set theory: {\em Union}, {\em Intersection}, and {\em Quorum} that will be described in Section~\ref{sec:method}.
With a heuristic function, a candidate feature set is identified at each iteration.
We compare the heuristic methods by demonstrating the anomaly detection performance when using the candidate feature set in the testing phase, with two public network traffic datasets (UNSW-NB15~\cite{moustafa2015unsw} and IDS2017~\cite{sharafaldin2018toward}).
Our experimental results show that {\em Intersection} and {\em Quorum} work better than   {\em Union} and produce  competitive subsets of features yielding the approximate performance to the one with the entire features. % for network anomaly detection. 
%the validity of the proposed method, yielding the approximate performance to the one with the entire features for network anomaly detection. 

This paper is organized as follows. Section~\ref{sec:bg} provides a short summary of the well-known techniques for variable selection  and the related studies.  
In section~\ref{sec:method}, we introduce our proposed method based on the ensemble approach, and the experimental results with the two public traffic datasets are demonstrated in Section~\ref{sec:eval}.
We finally conclude our presentation in Section~\ref{sec:conc} with a summary and future directions.

\section{Background}
\label{sec:bg}

This section offers the description of the feature selection classes and methods commonly applied in many application domains, and provides a summary of the closely related studies to our work.

\subsection{Feature selection classes}
%There are number of feature selection methods, and those feature selections are used for improving accuracy and reducing computation time. 
%Feature selection methods can be divided into roughly three categories, filter method, wrapper method, and embedded method. 
Feature selection is often used for improving accuracy and reducing computation time.
There are three categories for classifying feature selection methods: filter method, wrapper method, and embedded method. 

The filter method selects features based on intrinsic properties of data\cite{Noelia2007}. 
Using this method, the number of features can be reduced based on distance, consistency, similarity, and statistical measures~\cite{Jovic}. 
%This method depends on the general characteristics of training data to choose features~\cite{Noelia2007}.
There are two types in this class of methods: the univariate method  handles each feature independently, while the multivariate method  examines whole groups of features together.
The filter method is generally faster than the other classes of methods since it does not require a machine learning algorithm internally.
% performs better in generalization since it carry out  individually of the induction algorithm.
However, it has a tendency to choose a higher number of features; thus, it is usually used for pre-processing~\cite{Noelia2007}.

The wrapper method utilizes a machine learning classifier to measure the performance contribution provided by each feature. %s' performance and find out which features work well. 
Machine learning is also used to estimate each subset of features ~\cite{Jovic}.
Although the wrapper method is computationally expensive, it generally finds out a subset of features better than the one by  other methods.
%since it rely on the resource demands of classifier in each iteration.
However, the selected features could be biased depending on the classifier used, and a different learning method can be used for validating to reduce the risk of any potential bias.
%therefore, it is required to implement independent validation and another modeling algorithm to make sure if subsets are biased~\cite{Jovic}.

The embedded method also relies on a machine learning classifier. 
The embedded method uses a built-in classification algorithm for the feature search process; hence, the learning process and feature selection process are tightly coupled and not able to work individually.
This method is similar to the wrapper method, but it is generally faster since it does not examine all the possible combinations of features. 
Common methods in this class include decision tree, LASSO, and LARS~\cite{Dey2014}. 

\subsection{Feature selection methods}

Under the three classes of feature selection methods, there exists several specific methods widely used, and we considered the following methods in this study.

Correlation-Based Feature Selection (CFS) is a filter method which gives features ranking based on the heuristic evaluation. 
CFS seeks highly correlated and un-correlated features to discard redundant data.
Irrelevant features are disregarded since those features have low correlation.
Features highly correlated to one or more remaining features are considered as redundant features, and those features are screened out in this  selection scheme~\cite{Noelia2007}.

Sequential Backward Selection (SBS) is a wrapper method.
SBS uses a sequential  selection algorithm which is one of the greedy search algorithms, to reduce the number of features. 
SBS discards a feature one by one until it meets the number of features that a user provides as input. 
At first,  a criterion function should be defined to tell which feature is discarded Then the criterion function calculates the performance of after and before the elimination of each feature, and the feature showing the least performance  is discarded~\cite{book}.

Recursive Feature Eliminations (RFE) is also a wrapper method which relies on a machine learning classifier to discover the features least important.
RFE has been widely used for the cancer classification where the number of features is quite big (more than a thousand) and the training is done with less than 100 samples~\cite{chen2009rfe}.
%become one of the effective feature selection method~\cite{chen2009rfe}.
At each iteration, RFE drops one feature and re-rank remaining features, to remove redundant  and weak data through the process. 

%Gini Feature Importance uses Gini importance values from machine learning classifier. 
%And based on the importance values, features which have less than threshold will be removed from the subset. 
%Gini importance value is created when RF is constructing.
%When classification tree is building up, the attribute inducing the highest reduction of the Gini index will be selected at each node. 
%Assume \textit{p} as the fraction of associating pairs assign to node \textit{i} and \textit{1-p} as the fraction of non-associating pairs, Gini index is calculated as \textit{Gi = 2 p(1 − p)}~\cite{Tastan2009}.
The technique of Gini Feature Importance is one the embedded methods.
This scheme relies on a measure known as ``Importance value'', and a feature with a too small Importance value less than the specified threshold is removed from the feature set. 
The importance value is created when the internal random forest classifier is being constructed.
In detail, when a classification tree is built up, the variable inducing the highest reduction of the Gini index will be selected at each node. 
Gini index is calculated through $G_i = 2p(1-p)$~\cite{Tastan2009}, where
$p$ is a fraction of associating pairs assign to node $i$ and $(1-p)$ is a fraction of non-associating pairs.

%Univariate feature selection is one of the traditional feature selection. 
%It reduces dimensionality of features quick and efficacious so that this selection has been used long time.
%There are a lot of options for Univariate method including Pearson Correlation, Mutual information and maximal information coefficient (MIC), and Distance correlation.
%We implemented Chi-square test which investigate correlation or association of features using frequency distribution. 
Univariate feature selection belongs to the class of filter method, and reduces the dimensionality of features relatively quickly, which makes this scheme to be one of the traditional selection methods.
This technique evaluates each feature based on statistical tests and remove the variables having weak relationship with the output feature.
There are several options to utilize this method such as Pearson Correlation, mutual information and maximal information coefficient (MIC), and distance correlation.
We use this  scheme with the Chi-square test which calculates correlation or association of features based on the frequency distribution. 

\subsection{Related work}

There are a body of  studies investigated variable selection, and we introduce some of the previous studies closely related to our work.

%Parallel Variable Selection for Effective Performance Prediction
%The authors in~\cite{Wang2017}  proposed a parallel feature selection using Correlation CFS and SBS(sequential backward selection) to find best feature set. 
%This method is also using first and second round reductions to decrease features like our method. 
%It is similar to our method in the way, however, our method is not relying on single feature selection for second round reduction but also we are using multiple feature selections and heuristic approach to increase the accuracy. 
%%Furthermore, while this method can find best features after testing, our method is able to find best features during training period.
The authors in~\cite{Wang2017}  proposed a selection method based on  CFS and SBS to find a reduced feature set. 
However, the focus of this work is the parallelization of the selection scheme to minimize the time to build up a training model.
Hence, this work does not give an answer to when to stop to obtain a feature set that can actually be considered for training and testing, which is the research question we tackle to answer.

%A GA-LR wrapper approach for feature selection in network intrusion detection
Khammassi et al.~\cite{Khammassi2017} introduced a  wrapper approach which incorporates GA (Genetic Algorithm) feature search and LR (Logistic regression) for feature selection. 
This approach consists of three steps: pre-processing, feature selection, and  classification~\cite{Khammassi2017}.
The authors evaluated their proposed scheme with the UNSW-NB15 dataset, and reported 
the GA-LR method  yields 81.4\% of classification accuracy with 20 features.
As will be presented, our techniques outperforms showing over 86\% of accuracy only with 4 features selected.

%Analysis of network traffic features for anomaly detection
In the work of~\cite{Iglesias2015}, the authors provides an analysis of network traffic features for anomaly detection. 
The proposed selection method in this paper relies on feature weighting and ranking, such as WMR, SAM, and LASSO, to obtain a set of features with strong contribution. 
The selected features are then applied to a refinement process based on a brute force search analysis to decrease the number of features furthermore, by labeling relevant, medium relevant, and negligible. 
%This approach is similar to labeling Union, Quorum, and Intersection features. However, we are labeling not number of features, but feature subsets.
 
%Combining MIC feature selection and feature-based MSPCA for network traffic anomaly detection
Another recent work~\cite{Chen2017} proposed a detection system combining Maximal Information Coefficient (MIC) feature selection and feature-based MSPCA. 
%They used Maximal Information Coefficient which detects feature dependency between features and labels as a criteria, and Mic-Based Feature Selection Algorithm as a feature selection.
%MIC feature selection can remove uncorrelated features, however, using this method is tend to receive effects of outlier values, so it might discard important features.
MIC feature selection can remove uncorrelated features by detecting feature dependency between features and labels as a criteria and a feature selection algorithm called Mic-Based Feature Selection Algorithm.
A potential weakness of this method is the possibility to receive effects of outlier values, which might cause to discard an important feature.
\section{Proposed Method}
	\label{sec:method}

\begin{figure}[!tb]
\centering
\includegraphics[width=.90\columnwidth]{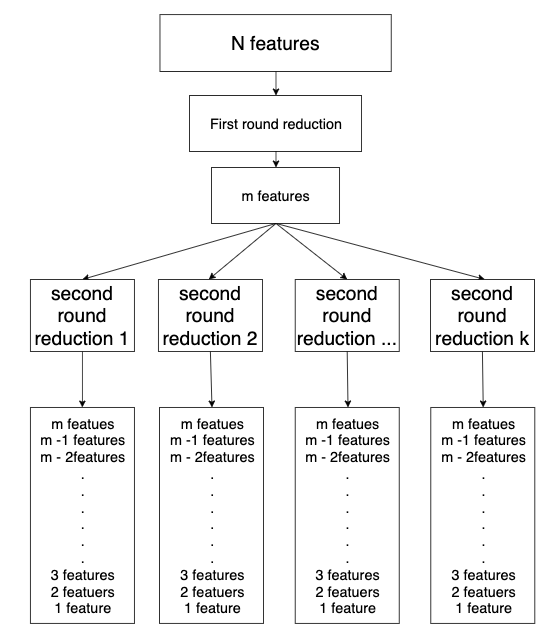}
\caption{
Proposed feature selection model consisting of two independent rounds:
The first-round  stage eliminates highly redundant features based on the pair-wise correlation factors, and the second-round stage employs a set of reduction methods, each of which  reduces the number of features based on an iterative process until it gets to a single feature. 
}
\label{fig:model1}
\end{figure}

%Our goal is to find an optimal subset of features during the training time, with which a classifier could approximate to the performance with the entire features.
%%with respect to classification performance with the reduced time complexity for training and testing. \fix{why is this a challenge?}
%To address this challenge, we take an ensemble approach by utilizing multiple selection methods, and the selection results discovered by the individual methods are combined to conclude a final result.

In this work, we utilize a set of selection techniques to identify a set of features that could be more important than the others with respect to the classification performance in the testing phase.
Fig.~\ref{fig:model1} 
%presents our ensemble-based selection method that consists of two reduction stages.
shows our proposed model composed of two independent rounds with multiple selection tools.
The first-round  stage eliminates highly redundant features based on the pair-wise correlation factors.
The second-round stage employs a set of reduction methods, each of which  reduces the number of features based on an iterative process until it gets to a single feature. 
As shown in the figure, the first-round selection reduces the number of features from $N$ to $m$ where $(m < N)$.
In the second-round process, multiple selection methods (from 1 to $k$ in the figure) eliminate one feature at each iteration independently; hence, $m$ iterations are required to complete.

\begin{figure}[!tb]
\centering
\includegraphics[width=.75\columnwidth]{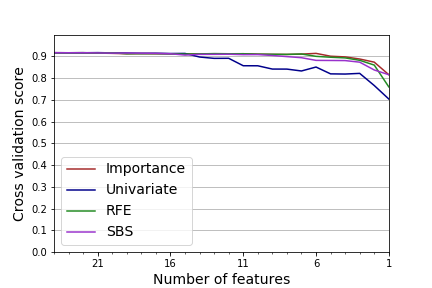}
\caption{
Second-round cross-validation result (dataset=UNSW-NB15):
the score is significantly dropped with the Univariate method  at around 11 features, while the other methods show a sharp degradation at three features.
}
\label{fig:unsw_cv_RF}
\end{figure}

Fig.~\ref{fig:unsw_cv_RF} shows an example with the training data file provided in the UNSW-NB15 dataset containing 39 numeric features.
The first-round process reduces from $N=39$ to $m=24$.
By default, we employ four selection methods for the second-round process: RFE, SBS, Univariate, and Importance. 
The reason why we chose these four methods is that these are common feature selection methods and  we would like to examine different classes of selection methods (i.e., filter, wrapper, and embedded) without a bias.
%In the second round, the four methods run independently and eliminate one feature at each iteration.
The figure shows the cross-validation results for each of $24$ iterations for the selection methods. 
From the figure, we can see that the score is significantly dropped with the Univariate method  at around 11 features, while the other methods show a sharp degradation at three features.
Since the variation selection is completed in the training phase, we may want to stop somewhere between round 11 ({\em conservatively}) and 3 ({\em optimistically}).

\begin{figure}[!tb]
 \centering
  \subfigure[Second-round testing result] {
    \label{fig:unsw_singleFS}
    \includegraphics[width=.75\columnwidth]{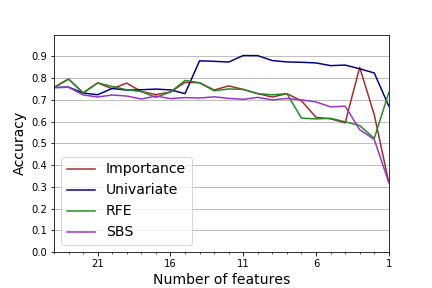}}
% \hspace{.1in}
 \subfigure[SBS testing result with different classifiers] {
    \label{fig:unsw_sbs_acc}
    \includegraphics[width=.75\columnwidth]{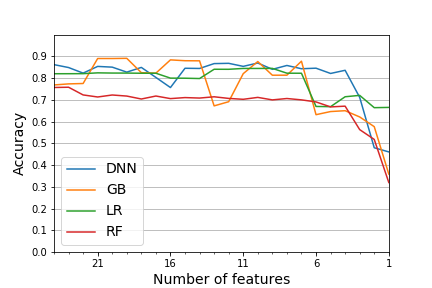}}
 \caption{
Testing accuracy (dataset=UNSW-NB15):
(a) Unlike the cross-validation result, the actual testing accuracy shows a degree of fluctuation over the number of features;
(b) SBS  works well with some  classifiers such as DNN and GB, while it works poorly with RF.
}
 \label{fig:unsw_testing}
\end{figure}

To see the quality of the selected features, we analyze the anomaly detection performance by using the testing data in UNSW-NB15. %, as shown in Fig.~\ref{fig:unsw_testing}.
Fig.~\ref{fig:unsw_singleFS} shows the actual testing result for the selection methods. 
We used the same classifier (RF) employed in Fig.~\ref{fig:unsw_cv_RF}.
Unlike  the  cross-validation result, the  testing accuracy shows a degree of fluctuation over the number of features.
Also surprisingly, Univariate shows a better result than the others, which is somewhat different from what we observed from the cross-validation result. 
%The figure shows that if we chose 3--13 features using Univariate, we may expect a higher accuracy in testing.
%As discussed however, it cannot be revealed in the training phase.
%
Fig.~\ref{fig:unsw_sbs_acc} shows the testing result when 
using  different classifiers of
%XGBoost (XGB), 
Logistic Regression (LR),
Gradient Boosting (GB), 
 Random Forest (RF) and Deep Neural Network (DNN), across the selected features by SBS.
In Fig.~\ref{fig:unsw_singleFS}, SBS did not show an impressive performance, but we can see that it may work well if we chose some other classifiers (e.g., DNN and GB), as shown in Fig.~\ref{fig:unsw_sbs_acc}.

An important observation here is that simply relying on the cross-validation result would be risky to determine which would work better and which does poorly.
In addition, a reduction method may work better with a certain classifier, and vice versa.
From these observations, we claim that simply choosing a single reduction technique may be risky.
Like SBS in Fig.~\ref{fig:unsw_sbs_acc},  other selection tools  produce their own sets of features at each iteration. 
Our proposed model incorporates the different selection methods to minimize the risk.
%
%As a result of the second round selection process, we have $k$ sets of features at each iteration, where $k$ is the number of selection methods employed.
One question here would then be {\em how to incorporate the individual results by different selection functions}. 
For this purpose, we define three heuristic methods based on set theory, as follows: 

\begin{itemize}
    \item {\em Union}, which takes all the features that any of the selection methods suggests;
    \item {\em Intersection}, which takes the features that everyone in the selection methods agrees upon;
    \item {\em Quorum}, which selectively takes the features agreed by the majority of the selection methods.
\end{itemize}

%We accomplished this goal by using common features from ensemble feature selections.  

\begin{figure}[!tb]
\centering
\includegraphics[width=.90\columnwidth]{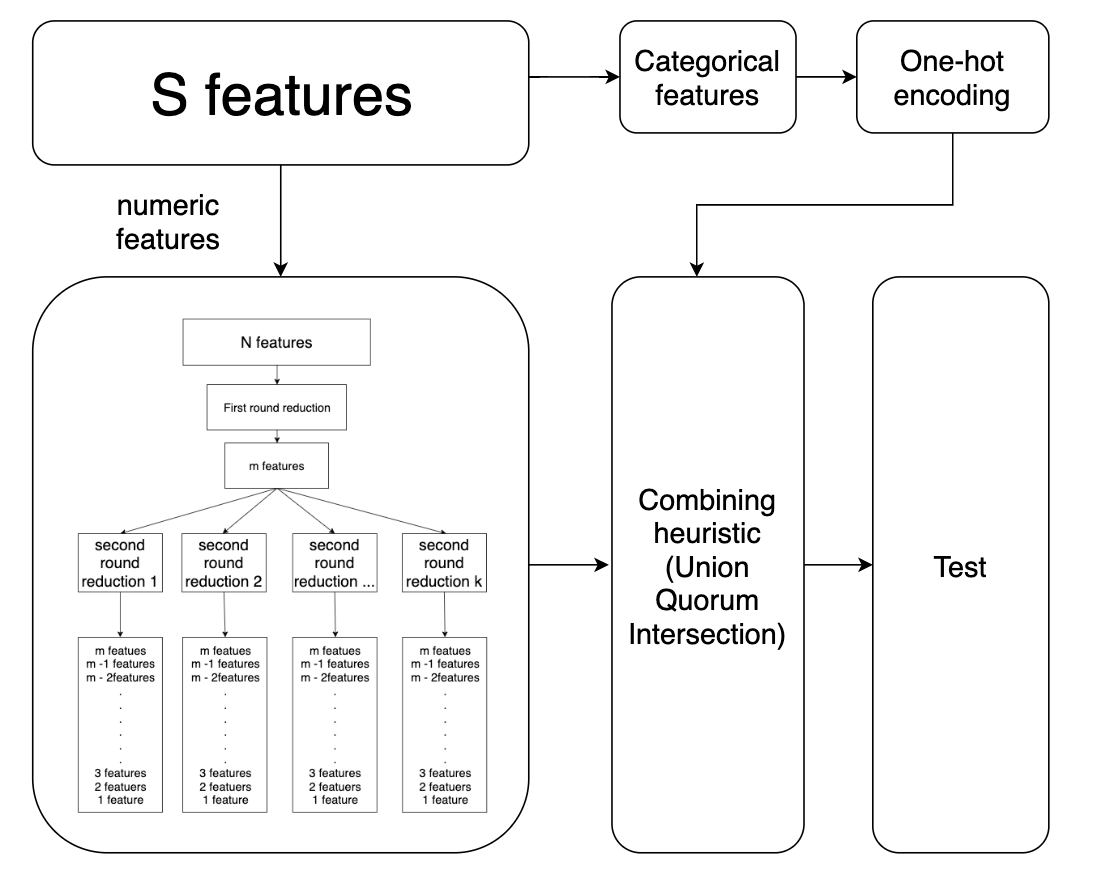}
\caption{
Ensemble feature selection: 
In the proposed selection process, the numeric features are fed into the ensemble engine, and the one-hot encoded features are then (optionally) added up to the final candidate features resulted by a heuristic method.
}
\label{fig:model2}
\end{figure}
   
Fig.~\ref{fig:model2} demonstrates the entire process of our feature selection method.
There typically exist two types of features: a numerical feature contains a numeric value (e.g., the number of bytes sent), while a categorical feature is discrete and % and has one out of the feature space set; 
an example of categorical features is {\tt protocol := \{TCP, UDP\}}.
A categorical feature is generally not supported by feature selection methods, and we employ one-hot encoding to convert it to $n$ numeric features with a binary value, where $n$ is the size of the feature space.
In the figure, the dataset consists of total $S$ features ($N$ numeric features and $S-N$ categorical features). 
The numeric features are fed into the ensemble engine in Fig.~\ref{fig:model1}.
The one-hot encoded features are then (optionally) added up to the final candidate features resulted by a heuristic method.

%However, categorical features are very important features in some dataset. 
%Therefore, we handled categorical features separately. 
%While numeric features go to the figure 1 directly,categorical features are applied one-hot encoding. 
%When it comes to cross-validation section, those categorical features are merged.
%After that, Union, Majority, Intersection are gathered from the ensemble feature selections and these features are predicting test accuracy.

%As figure# shown, our method utilize two rounds of reduction. 
%The first round reduction eliminates similar data by correlation filter method in order to get rid of redundant data.  
%In the second round, each feature selection method reduce number of features one by one.     

%We implemented ransom forest using 100 trees from Python scikit-learn library. 

\section{Experiments}
    \label{sec:eval}
    
In this section, we report our experimental results conducted with the UNSW-NB15 and IDS2017 datasets.
%As in the previous section, the four selection methods are considered for the second-round reduction.
Note that the samples in the datasets are pre-processed including the normalization before being used.
%In our implementation, we used 100 trees for Random Forest, 200 trees for Gradient Boosting, and A Library for Large Linear Classification solver for Logistic Regression. 
%For Deep Neural Network, we used 4 hidden layers which contains 50 hidden units with a rectified linear unit (ReLU) activation function and dropout of 0.2 in each layer. 
%The output layer contains 1 unit with sigmoid activation function. 
%The optimizer is Adam optimization algorithm, and the batch size is 100 with 15 epochs.
For the classifiers, we basically configured them with the default setting and here are some specifics: 100 trees for RF, 200 trees for GB, and a library for Large Linear Classification solver for LR. 
For DNN, we used 4 hidden layers which contains 50 hidden units with a rectified linear unit (ReLU) activation function and the dropout of 0.2 in each layer. 
The output layer contains 1 unit with Sigmoid activation function. 
We used an Adam optimizer, and the batch size is 100 with 15 epochs.

\subsection{Evaluation metrics}

We basically examine the classification performance by utilizing multiple classifiers, including LR, GB, RF, and DNN.
For reporting performance, we utilize {\em F1-score} that is a harmonic mean of {\em Precision} and {\em Recall}, defined by the elements in the confusion matrix: TP (True Positive), FP (False Positive), FN (False Negative) and TN (True Negative). 
Since the metric of {\em Accuracy} may lead to a biased result if the population of the minority class is too small, F1-score is widely accepted to minimize the concern; for this reason, we use the measure of F1-score to measure the performance of classifiers.
The metrics of Accuracy and F1-score are  defined:

\iffalse
\begin{itemize}
   \item Accuracy = $\frac{TP+TN}{TP+TN+FP+FN}$
   \item Precision = $\frac{TP}{TP+FP}$
   \item Recall = $\frac{TP}{TP+FN}$
   \item F1-score = $2 \cdot \frac{\text{Precision} \times \text{Recall}}{\text{Precision} + \text{Recall}}$
\end{itemize}
\fi
{\small
$$\text{Accuracy} = \frac{TP+TN}{TP+TN+FP+FN}$$
%$$\text{Precision} = \frac{TP}{TP+FP}$$
%$$\text{Recall} = \frac{TP}{TP+FN}$$
%$$\text{F1-score} = 2 \cdot \frac{\text{Precision} \times \text{Recall}}{\text{Precision} + \text{Recall}}$$
$$\text{F1-score} =  \frac{2 
\times \frac{TP}{TP+FP} \times \frac{TP}{TP+FN}}{\frac{TP}{TP+FP} + \frac{TP}{TP+FN}}$$
}
 
\subsection{Experimental result for UNSW-NB15}

\begin{figure}[!tb]
 \centering
 \subfigure[Iterations] {
    \label{fig:unswRF_round}
    \includegraphics[width=.75\columnwidth]{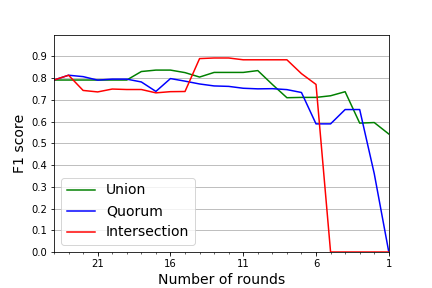}}
% \hspace{.1in}
 \subfigure[Number of features] {
    \label{fig:unswRF_feature}
    \includegraphics[width=.75\columnwidth]{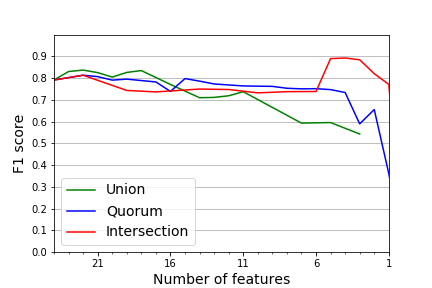}}
 \caption{
Performance of heuristics (classifier=RF, dataset=UNSW-NB15): 
The candidate size showing the best performance for {\em Intersection} is ranged from 5--3 features, which is a competitive result to the use of the full feature set (89.6\% in F1-score). 
{\em Union} seems to produce less competitive feature sets along the iterative process, showing lower than 70\% in F1-score with 10 or  smaller number of features.
The method of {\em Quorum} shows the best performance for the feature set size from 15 to 6, but is worse than {\em Intersection} with less than six features.
 }
 \label{fig:unsw_RF}
\end{figure}

\begin{figure}[!tb]
 \centering
  \subfigure[LR] {
    \label{fig:unswLR}
    \includegraphics[width=.75\columnwidth]{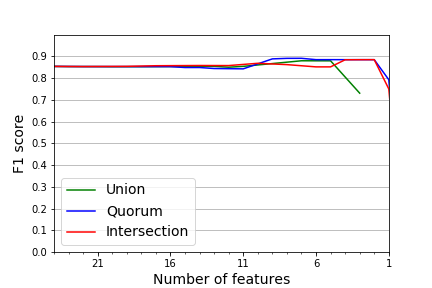}}
% \hspace{.1in}
 \subfigure[GB] {
    \label{fig:unswGB}
    \includegraphics[width=.75\columnwidth]{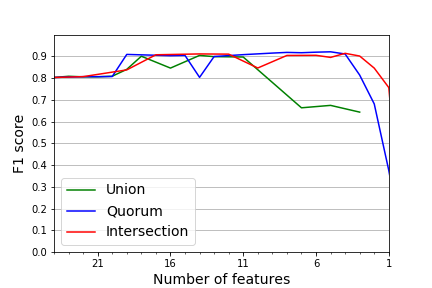}}
% \hspace{.1in}
% \subfigure[RF] {
%    \label{fig:unswRF}
%    \includegraphics[width=.75\columnwidth]{unswRF.png}}
%% \hspace{.1in}
  \subfigure[DNN] {
    \label{fig:unswANN}
    \includegraphics[width=.75\columnwidth]{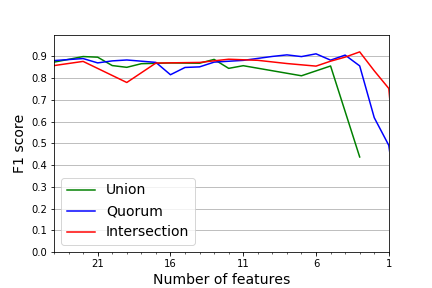}}
 \caption{
Performance of heuristics with different classifiers (dataset=UNSW-NB15):
(a) LR: {\em Quorum} and {\em Intersection} result in almost 90\% in F1-score, which is higher than the performance with the entire numeric features (85.2\%);
(b) GB: {\em Quorum} and {\em Intersection} are able to discover competitive feature sets, approaching to the performance with the entire features (90.9\%) when using the GB classifier; 
(c) DNN: {\em Quorum} and {\em Intersection} produce the approximated performance to the full feature set (90.0\%)  with only 3--4 features.
}
 \label{fig:unsw_classifiers}
\end{figure}

\iffalse
\begin{figure}[!tb]
 \centering
 \subfigure[Numeric features only] {
    \label{fig:unswRF_round}
    \includegraphics[width=.45\columnwidth]{unswRF_round.png}}
% \hspace{.1in}
 \subfigure[Including one-hot encoded features] {
    \label{fig:unswRF_round_ohe}
    \includegraphics[width=.45\columnwidth]{unswRF_round_ohe.png}}
 \caption{

 }
 \label{fig:unsw_classifiers}
\end{figure}
\fi
\begin{figure}[!tb]
\centering
\includegraphics[width=0.75\columnwidth]{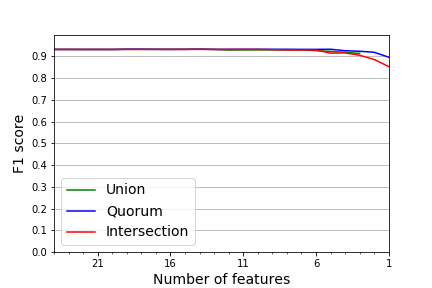}
\caption{
Performance of heuristics with one-hot encoded features (classifier=RF, dataset=UNSW-NB15): 
Choosing 5+ features with any heuristic allows to approximate to the performance with the full feature set.
}
\label{fig:unswRF_ohe}
\end{figure}

We employ the raw UNSW-NB15 dataset collected in a simulated environment %with servers and routers 
in January 22, 2015  and February 17, 2015. 
%IXIA traffic generator was designed by three virtual servers. 
%Two servers are used for creating normal network and one server is for creating for malicious network traffic. 
%Servers and IXIA were connected with two virtual interface to attach  private and public network. 
%And servers were connected by hosts through two routers. Routers were linked to Firewall to find out traffics which were not detected by traditional signature-based tool. 
%Total size of UNSW-NB15 is approximately 100GB which recorded 2 million observations.
The dataset provides separate files for training and testing data.
The training file contains  roughly 82K data points and 55\% of them are anomalies.
The testing file includes 175K samples and 68\% of the data points are anomalies. 
The anomalies are mainly by cyber-attacks, and nine classes of attacks were injected to collect.
There are 42 features available excluding the serial number and label information, and 39 of them are numeric.

As a result of our ensemble method, % (shown in Fig.~\ref{fig:model1}), 
we obtain the cross validation score as shown in Fig.~\ref{fig:unsw_cv_RF}.
Recall that  cross validation takes place only with the training file, while the measured F1-score in this section is for actual testing with an independent testing file.
Fig.~\ref{fig:unsw_RF} shows how the proposed heuristic methods execute  across the iterations (Fig.~\ref{fig:unswRF_round}) and  the number of features (Fig.~\ref{fig:unswRF_feature}), when using the RF classifier.
At each round (iteration), our proposed methods combine the common features identified by the second-round reduction functions.
Intuitively, {\em Union} produces a larger number of features while {\em Intersection} has the smallest number of features at a certain round.
As shown in the figure, {\em Intersection} identifies good candidate sets approaching 90\% in F1-score, at round 14--8.
By referencing Fig.~\ref{fig:unswRF_feature}, we can see that the candidate size showing the best performance for {\em Intersection} is ranged from 5--3 features, which is a competitive result to the use of the full feature set (89.6\% in F1-score). 
From the figure, {\em Union} seems to produce less competitive feature sets along the iterative process, showing lower than 70\% in F1-score with 10 or  smaller number of features.
The method of {\em Quorum} shows the best performance for the feature set size from 15 to 6, but is worse than {\em Intersection} with less than six features.

Considering different classifiers along the number of features,  Fig.~\ref{fig:unsw_classifiers} shows the result of the heuristic methods.
In~\ref{fig:unswLR}, LR results in almost flat curves over the number of features, which implies that it is possible to significantly reduce the feature size for training and testing with a minor penalty in performance. 
The methods of {\em Quorum} and {\em Intersection} result in almost 90\% in F1-score, which is higher than the performance with the entire numeric features (85.2\%).
Fig.~\ref{fig:unswGB} shows {\em Quorum} and {\em Intersection} are able to discover competitive feature sets, approaching to the performance with the entire features (90.9\%) when using the GB classifier. 
The deep learning classifier (DNN) in Fig.~\ref{fig:unswANN} makes the similar conclusion, and {\em Quorum} and {\em Intersection} produce the approximated performance to the full feature set (90.0\%)  with only 3--4 features.

In the discussion of Fig.~\ref{fig:model2}, we mentioned that the categorical features could be added to the incorporation process by heuristics.
Fig.~\ref{fig:unswRF_ohe} shows the performance of the heuristic methods with the addition of three one-hot encoded features defined in UNSW-NB15.
In the figure, much consistent results are shown with the heuristics, and choosing five or more features with any heuristic allows to approximate to the performance with the full feature set.

\subsection{Experimental results for IDS2017}
\begin{figure}[!tb]
 \centering
\vspace{-10pt}
  \subfigure[Cross validation] {
    \label{fig:ids_CV}
    \includegraphics[width=.75\columnwidth]{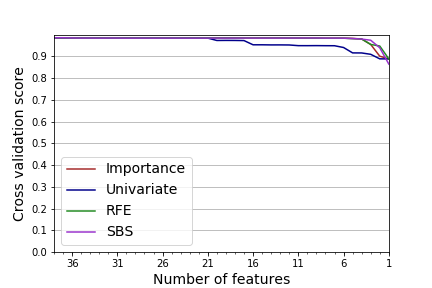}
    \vspace{-20in}}
\subfigure[RF] {
    \label{fig:ids_RF}
    \includegraphics[width=.75\columnwidth]{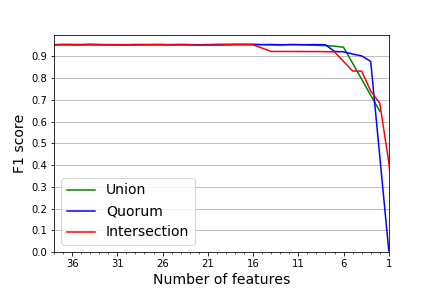}
    \vspace{-20pt}}
% \hspace{.1in}
 \subfigure[DNN] {
    \label{fig:ids_ANN}
    \includegraphics[width=.75\columnwidth]{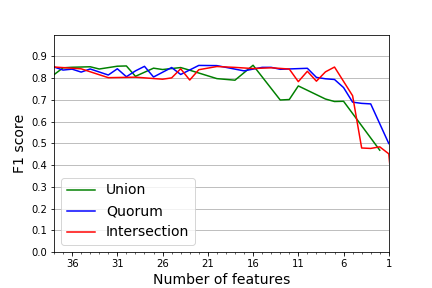}}
 \caption{
Experimental result (dataset=IDS2017):
(a) Cross validation result shows a quite high score over 90\% even with a single feature; 
(b) \& (c): {\em Quorum} and {\em Intersection} result in higher quality features in the reduction.
 }
 \label{fig:ids_eval1}
\end{figure}

We next share our observations from the experiments with the IDS2017 dataset.
This dataset was collected over the five days from July 3, 2017 to July 7, 2017. 
Similar to UNSW-NB15, it was collected in a simulated network environment with the victim network and attackers network. 
IDS2017 includes many kinds of attacks, including FTP, SSH, DoS/DDoS, Heartbleed, Infiltration, Botnet, and Web attacks. 
%Also,  25 users based on the HTTP, HTTPS, FTP, SSH, and email protocols.
%In this dataset, a certain day trace contains no or specific attacks only. \fix{fix this previous sentence, "in this dataset..."}
In this dataset, daily trace records include specific attacks selectively.
For this reason, we randomly sampled 100K data points from the entire records, one sampling for training and the other for testing.
The total number of features in this dataset is 84, and we excluded the first five features related to the source and destination information.
All 79 features used in our experiments are numeric.

Fig.~\ref{fig:ids_eval1} shows the experimental result with the IDS2017 dataset.
The cross validation result in Fig.~\ref{fig:ids_CV} shows a quite high score over 90\% even with a single feature. 
Fig.~\ref{fig:ids_RF} and Fig.~\ref{fig:ids_ANN} show how the two classifiers of RF and DNN work across the number of features.
From the result, we can see RF works better than DNN for this dataset.
Overall, {\em Quorum} and {\em Intersection} result in higher quality features in the reduction.
With RF, using five features is sufficient to approximate to the one with the entire features, while using 7 or more features would be desirable for DNN.

Although omitted due to the space reason, we conducted the same experiment with other pairs of training and testing files sampled with different seed numbers, and observed similar patterns to the result reported in Fig.~\ref{fig:ids_eval1}.

\section{Conclusion}
	\label{sec:conc}
Variable selection is beneficial by reducing the training and testing complexity, but it generally requires highly laborious efforts with intensive analysis to identify an optimal subset of features in advance.
Automating the selection process is thus essential for the practical use of variable selection, but it is challenging with the existing methods since it is hard to determine the termination condition to stop searching. In addition, we observed that individual selection methods result in different feature sets as the output. 
%This study shows results toward enabling automated variable selection, and
In this paper, we  presented an ensemble method that combines the independent results by individual selection methods through heuristic functions, in order to identify an optimal set of features.
%we defined  three heuristics: (1) {\em intersection} taking the features everyone agrees upon as a candidate feature set, (2) {\em union} taking the ones anyone suggests, and (3) {\em quorum} selecting ones a majority of the participants (selection methods) consent on.
Our experimental results showed that the heuristic functions of {\em Intersection} and {\em Quorum} work consistently producing competitive subsets of features with the approximate performance to the one with the entire features for network anomaly detection. 
The next step of this research is to develop a method that helps determine when to stop in the iterative process in an automated fashion without human intervention.

\section*{Acknowledgment}
This effort was supported in part by the U.S. Department of Energy (DOE), Office
of Science, Office of Advanced Scientific Computing Research under
contract number DE-AC02-05CH11231, and in part by Institute for Information \& communications Technology Promotion (IITP) grant funded by the Korea government (MSIP) (No.2016-0-00078, Cloud-based Security Intelligence Technology Development for the Customized Security Service Provisioning).
This research used resources of the National Energy Research Scientific Computing Center (NERSC), a DOE Office of Science User Facility.

% ADD THE FOLLOWING COUPLE LINES INTO YOUR PREAMBLE
\let\OLDthebibliography\thebibliography
\renewcommand\thebibliography[1]{
  \OLDthebibliography{#1}
  \setlength{\parskip}{0pt}
  \setlength{\itemsep}{2pt plus 0.3ex}
}

\bibliographystyle{unsrt}
\bibliography{ref2}

\end{document}